\documentclass[10pt,twocolumn,letterpaper]{article}

\usepackage{cvpr}              

\usepackage[dvipsnames]{xcolor}

\usepackage{bm}
\usepackage[abs]{overpic}
\AtBeginDocument{%
  \setlength\abovedisplayskip{2pt}
  \setlength\belowdisplayskip{2pt}}
\allowdisplaybreaks

\def\sysname{PI3D\xspace}

\newlength\paramargin
\newlength\figmargin

\newlength\secmargin
\newlength\figcapmargin
\newlength\tabcapmargin

\newif\ifReview\Reviewfalse

\newcommand{\lfi}[1]{\textbf{#1}}

\long\def\ignorethis#1{k}

\newbox\jsavebox%

\makeatletter
\newcommand{\providelength}[1]{%
  \@ifundefined{\expandafter\@gobble\string#1}
   {
    \typeout{\string\providelength: making new length \string#1}%
    \newlength{#1}%
   }
   {
    \sdaau@checkforlength{#1}%
   }%
}

\definecolor{cvprblue}{rgb}{0.21,0.49,0.74}
\usepackage[pagebackref,breaklinks,colorlinks,citecolor=cvprblue]{hyperref}
\usepackage[accsupp]{axessibility}
\usepackage{bbding}
\usepackage{amssymb}
\usepackage[normalem]{ulem}
\usepackage{multirow}
\usepackage{multicol}
\usepackage{graphicx}
\usepackage{caption}

\def\paperID{1715} 
\def\confName{CVPR}
\def\confYear{2024}

\title{PI3D: Efficient Text-to-3D Generation with Pseudo-Image Diffusion}

\author{
Ying-Tian Liu$^{1}$\quad
Yuan-Chen Guo$^{1,2}$\footnotemark[1]\quad
Guan Luo$^{1}$ \quad
Heyi Sun$^{1}$ \quad
Wei Yin$^{4}$ \quad
Song-Hai Zhang$^{3,1}$\footnotemark[2]\\
$^{1}$ BNRist, Tsinghua University\;
$^{2}$ VAST\;
$^{3}$ Qinghai University \;
$^{4}$ The University of Adelaide
\\
}

\begin{document}
\twocolumn[{%
\renewcommand\twocolumn[1][]{#1}%
\maketitle
\begin{center}
    \centering
    \captionsetup{type=figure}
    \vspace{-8mm}
    \includegraphics[width=\linewidth]{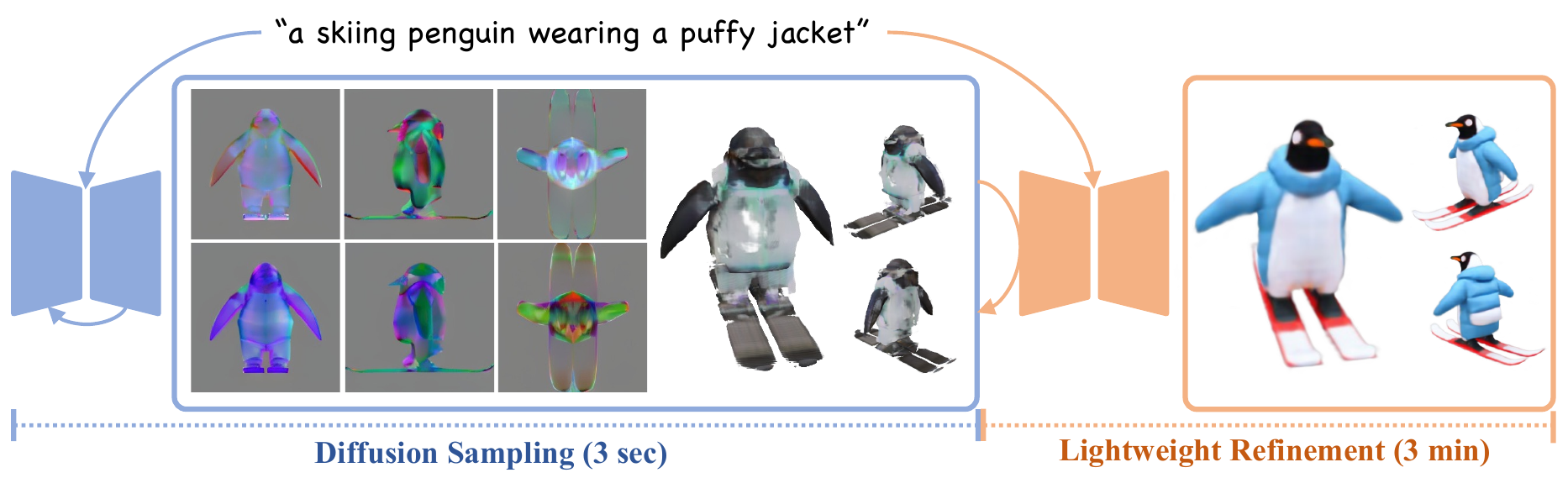}
    \captionof{figure}{Our method, \sysname, is able to generate the pseudo-images of a 3D shape in seconds from text prompts and further refine it in 3 minutes to achieve better quality.}
    \label{fig:teaser}
\end{center}%
}]

\renewcommand{\thefootnote}{\fnsymbol{footnote}}
\footnotetext[1]{ Work done during internship at VAST.}
\footnotetext[2]{ Corresponding author.}

\begin{abstract}

Diffusion models trained on large-scale text-image datasets have demonstrated a strong capability of controllable high-quality image generation from arbitrary text prompts. However, the generation quality and generalization ability of 3D diffusion models is hindered by the scarcity of high-quality and large-scale 3D datasets. In this paper, we present \sysname, a framework that fully leverages the pre-trained text-to-image diffusion models' ability to generate high-quality 3D shapes from text prompts in minutes. The core idea is to connect the 2D and 3D domains by representing a 3D shape as a set of \textbf{P}seudo RGB \textbf{I}mages. We fine-tune an existing text-to-image diffusion model to produce such pseudo-images using a small number of text-3D pairs. Surprisingly, we find that it can already generate meaningful and consistent 3D shapes given complex text descriptions. We further take the generated shapes as the starting point for a lightweight iterative refinement using score distillation sampling to achieve high-quality generation under a low budget. \sysname generates a single 3D shape from text in only 3 minutes and the quality is validated to outperform existing 3D generative models by a large margin.

\end{abstract}    
\section{Introduction}
\label{sec:intro}
With recent advances in generative modeling, various models have emerged capable of capturing complex data distributions across different modalities such as audio, video, and images. Notably, pioneering image generation models like Imagen~\cite{imagen}, DALL-E~\cite{dalle,dalle2}, Stable Diffusion~\cite{ldm}, and others demonstrate the ability to generate highly realistic and natural-looking images from text prompts, often indistinguishable from those produced by humans.

While algorithmic progress plays a crucial role in these achievements, it's essential to acknowledge the foundational role of large-scale datasets~\cite{laion5B} in facilitating such applications. Unlike 2D images, 3D objects face inherent challenges in data acquisition. The labor-intensive and time-consuming processes involved in 3D scanning and creation make gathering 3D data notably challenging. Although datasets like Objaverse~\cite{objaverse} and Objaverse-XL~\cite{objaverse-xl} have recently been released with millions of 3D assets, they suffer from insufficient filtering and processing, leading to the inclusion of duplicate, low-quality, and texture-less content. The absence of a high-quality, abundant, and semantically rich 3D dataset remains a significant obstacle in directly constructing 3D generative models.

In order to generate 3D assets in scenarios where 3D priors are difficult to obtain, researchers have recently been keen to utilize various 3D representations~\cite{nerf,neus,gsplat,dmtet} to lift 2D diffusion models with Score Distillation Sampling (SDS)~\cite{dreamfusion, sjc}. However, applying only 2D prior is prone to hinder the 3D nature, inevitably compromising 3D fidelity and consistency. 3D knowledge is a key issue that cannot be bypassed for high-quality 3D generation.

The pivotal question arises: can a 3D diffusion model exhibit robust generalization when 3D data cannot reach the scale and quality of image data? We answer this question by presenting \sysname, a framework that fully utilizes the knowledge in pre-trained text-to-image diffusion models and image datasets to generate high-quality 3D shapes conditioned on text prompts. 
\begin{figure}[t]
    \centering
    \scalebox{0.9}{
     \begin{overpic}[scale=0.4]{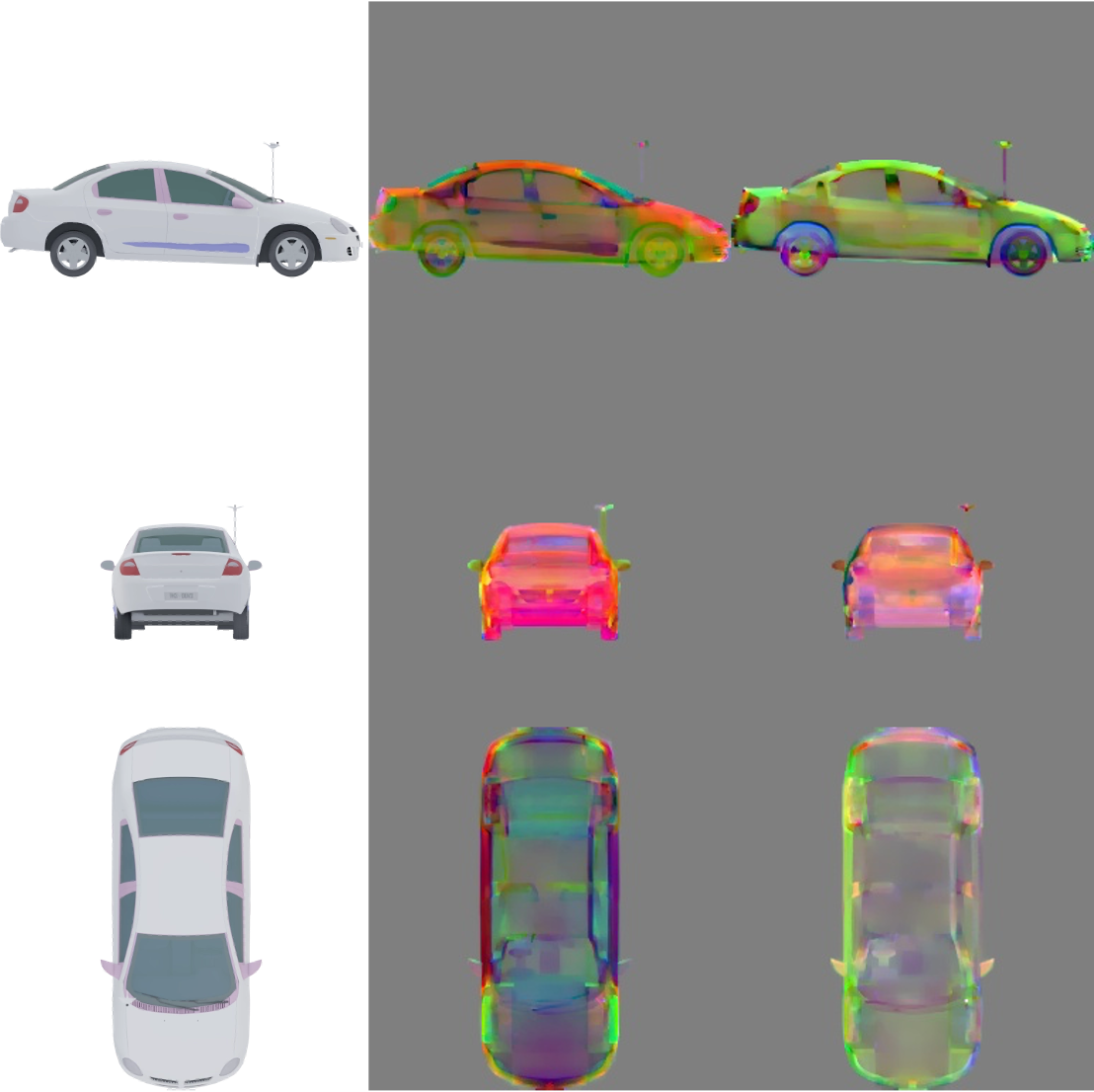}
     \put(0, -15){Orthogonal Views}
     \put(110, -15){Pseudo-Images}
    \end{overpic}
    }
    \vspace{3mm}
    \caption{We show the 3 orthogonal rendering results and the triplane representation as 6 pseudo-images for the same scene. We can observe semantic congruence between them, such as the contours of different parts.}
    \vspace{-8mm}
    \label{fig:triplane-image}
\end{figure}
Given the observation that the triplane representation for 3D scenes~\cite{convolutional_occupancy_networks,eg3d} shares semantic congruence with orthogonal rendered images, we believe that the architecture of a pre-trained image diffusion model is also capable of generating high-quality triplanes. In order to borrow knowledge from 2D generation, we treat every 3 channels of the triplane representation as one image to form a set of pseudo RGB images (see \cref{fig:triplane-image}). To build our 3D diffusion model, we start from a pre-trained text-to-image diffusion model and keep most of its architecture to inherit the generalizability while changing the self-attention layer to global attention performed on all pseudo-images. We inject 3D knowledge into the pre-trained 2D diffusion model by fine-tuning it on the pseudo-images obtained from 20k text-3D pairs from the Objaverse dataset~\cite{objaverse}. Further, we find that the simultaneous use of 2D image data in training significantly improves generalization ability on complex concepts while keeping the requirements of the magnitude of 3D data at a low level. Based on the pseudo-image diffusion model, we can create brand-new 3D objects from text prompts by performing diffusion sampling in seconds. We take the sampled object as the starting point for a subsequent refinement process where score distillation sampling (SDS)~\cite{dreamfusion} is used to further improve the quality. The refinement process converges quickly and maintains good 3D consistency thanks to the strong initialization provided by the pseudo-image diffusion model. \sysname is able to generate high-quality 3D objects from text prompts within 3 minutes, bringing new possibilities for 3D content creation.

The main contributions of this paper are as follows:
\begin{itemize}
    \item We propose \sysname, a framework that fully leverages prior knowledge from 2D diffusion models for high-quality and text-aligned 3D generation in minutes.

    \item We propose to represent a 3D shape as a set of pseudo-images and adapt a 2D diffusion model to directly output such a representation, enabling fast sampling of 3D objects from texts while providing strong initialization for further refinement.

    \item \sysname is demonstrated to be effective in 3D generation with respect to visual quality, 3D consistency, and generation speed.
    
\end{itemize}

\section{Related Work}
In this section, we will primarily review recent text-to-3D research. They can be broadly categorized based on the data they utilize and how they incorporate 3D knowledge: methods using 3D diffusion methods and lifting 2D diffusion models.

\subsection{Diffusion-based 3D Generative Model}
This line of research focuses on training 3D diffusion models on popular 3D datasets~\cite{shapenet,objaverse,objaverse-xl},  each with distinct representations of 3D objects. 
As most of these methods have access to 3D raw data, they inherently acquire 3D knowledge inductively from the dataset. Point·E~\cite{pointe} trains two point-cloud diffusion models conditioned on the text prompt or single-view image to generate a low-resolution point cloud and up-sample it respectively. It can further obtain meshes from the point cloud via SDF prediction. Shap·E~\cite{shape} trains the diffusion model on the 3D implicit function. It is capable of simultaneously producing NeRF and SDF with two heads, but it struggles with semantics when faced with complex prompts. NFD~\cite{nfd}, 3DGen~\cite{3dgen}, and DiffTF~\cite{large-vocabulary} adopt diffusion on the triplane representation to achieve highly efficient mesh generation but they constrain the generation to be category-specific on ShapeNet~\cite{shapenet} and OmniObject3D~\cite{omniobject}. Some other methods~\cite{holodiffusion,holofusion} build 3D diffusion models with only posed 2D images on smaller datasets~\cite{co3d}. All these methods actually utilize only poorly labeled 3D data or limited image data, which leads to their poor ability to generate reasonable objects when confronted with some complex concepts.


\subsection{3D Knowledge in Lifting 2D Diffusion Models}
Several recent works focus on optimizing a 3D representation by leveraging a pre-trained 2D diffusion model. The foundational technique involved is Score Distillation Sampling (SDS)~\cite{dreamfusion,sjc}. SDS aims to align the rendering results of a parameterized renderer with the distribution described by a pre-trained image diffusion model, achieving parameter space sampling. In DreamFusion~\cite{dreamfusion} and SJC~\cite{sjc}, 3D knowledge originates solely from spatial representations and the view-dependency prompting, which are relatively weak. This limitation increases the likelihood of 3D inconsistency, such as the Janus problem, since there is no direct informative constraint between different views. ProlificDreamer~\cite{vsd} and HiFA~\cite{hifa} reduce the oversaturation and greatly increase the generation quality by reinventing the optimization target. Perp-Neg~\cite{perpneg} mitigates the inconsistency problem by correcting the direction of parameter updates guided by the 2D diffusion model but leaves the lack of a 3D prior unsolved. Zero-1-to-3~\cite{zero123} proposes to train a 2D viewpoint transformation model conditioned on the input view and the relative camera transformation, injecting implicit 3D knowledge into the 2D diffusion model. Magic123~\cite{magic123} fuses and balances guidance from pre-trained 2D diffusion model~\cite{ldm} and viewpoint transformation model~\cite{zero123}, achieving detailed image-to-3D results. MVDream~\cite{mvdream} models the distribution of multi-view images and thus greatly improves the stability of existing 2D-lifting methods. ATT3D~\cite{att3d} amortizes the optimization cost among prompts by introducing a prompt-conditioned parameter prediction hyper-network. DreamGaussian~\cite{dreamgaussian} accelerates the generation by utilizing real-time rendered 3D representation. Although these methods are usually capable of generating an attractive appearance, they all require tons of optimizing iterations and are of lower diversity. 

Some other methods view 3D generation as multi-view generation combined with reconstruction~\cite{instant3d,syncdreamer,mvdiffusion,one2345}. The reconstruction process can be optionally replaced by a feed-forward model~\cite{one2345, lrm}. These approaches allow for fast 3D generation but are often jointly affected by minor inconsistencies of multi-view images and errors in the reconstruction process.




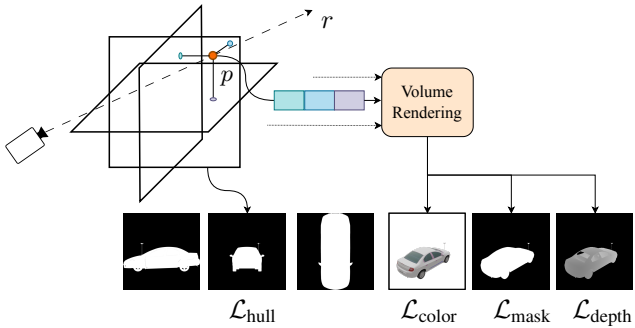
\begin{figure}[htbp]
    \centering
    \begin{overpic}[scale=0.4]{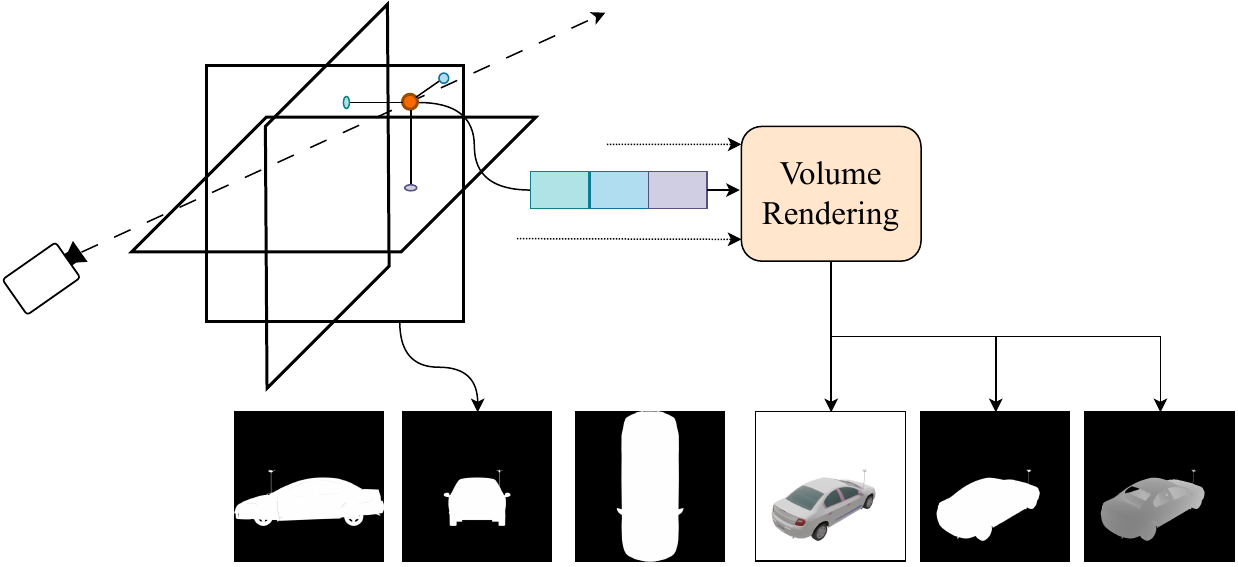}
    \put(120,100){$r$}
    \put(82,80){\small $p$}
    \put(85,-10){$\mathcal{L}_{\text{hull}}$}
    \put(150,-10){$\mathcal{L}_{\text{color}}$}
    \put(185,-10){$\mathcal{L}_{\text{mask}}$}
    \put(215,-10){$\mathcal{L}_{\text{depth}}$}
    \end{overpic}
    \vspace{1px}
    \caption{We fit the triplane representation for each object with the supervision of the rendered RGB images, binary mask, and object depth. We also adopt axis-aligned masks on the triplane to encourage the geometry to approach the correct surface faster. }
    \vspace{-4mm}
    \label{fig:triplane}
\end{figure}

\section{Method}
In this section, we mainly focus on how \sysname is built. It consists of two parts as a whole, the first part is a pseudo-image diffusion model that is responsible for fast sampling of 3D objects. It follows two stages similar to many other latent diffusion methods~\cite{3dgen,neuralwave,sdfusion,shape}. In the first stage (in \cref{sec:triplane-fit}), a 3D representation for each 3D object is fitted and next (in \cref{sec:pid}), we build a text-conditioned pseudo-image diffusion model adapting from a pre-trained 2D diffusion model. The second part (in \cref{sec:refine}) is a lightweight refinement using Score Distillation Sampling (SDS) that converges quickly and further improves the quality of the sampled 3D object. 

\subsection{Depth-Aware Triplane Fitting}
\label{sec:triplane-fit}
We represent each object as three axis-aligned feature maps $F_{xy}, F_{xz}, F_{yz} \in \mathbb{R}^{C\times H\times W}$. Specifically, we set $C = 6$ to form 6 pseudo-images. The spatial feature $F(p)$ for any query point $p$ is constructed by concatenating the features bi-linearly interpolated on the three maps at the projected 2D positions.
\begin{equation}
    F(p) = [F_{xy}(p_{xy}), F_{xz}(p_{xz}), F_{yz}(p_{yz})]
\end{equation}
The color and density at the point $p$ are then decoded by a shared MLP $\mathcal{D}(\cdot;\theta)$.
\begin{equation}
    c(p), \sigma(p) = \mathcal{D}(F(p);\theta)
\end{equation}
The expected color of a camera ray $r$ is calculated in a volume rendering manner~\cite{volume_rendering,nerf}. 
\begin{equation}
\begin{aligned}
    \hat{C}(r) = & \sum_{i=1}^N T_i (1-\exp(-\delta\sigma(p_i)))c(p_i), \\ & \text{where}\ T_i = \exp(-\sum_{j=1}^{i-1}\delta\sigma(p_j))
\end{aligned}
\end{equation}
$\{p_i\}_{i=1}^N$ denotes the sampled points along that ray. We do not involve additional positional embedding like in vanilla NeRF~\cite{nerf}, since we want the decoder to be position-agnostic, ensuring that what is similar on the rendered images is also similar on the triplane. 
To encourage a correct underlying geometry, we also supervise the object mask $\hat{M}(r)$ and the expected depth $\hat{D}(r)$ along that ray, which are given by: 
\begin{equation}
    \hat{M}(r)= \sum_{i=1}^N T_i (1-\exp(-\delta\sigma(p_i))) 
\end{equation}
\begin{equation}
   \hat{D}(r) = \sum_{i=1}^N T_i (1-\exp(-\delta\sigma(p_i)))t(p_i)
\end{equation}
where $t(p)$ is the distance from the ray origin to $p$. 

The losses on rendering views are as follows:
\begin{equation}
        \mathcal{L}_{\text{color}} = \sum_r (||C(r) - \hat{C}(r)||_2 + \lambda_{\text{L1}}||C(r) - \hat{C}(r)||_1) 
\end{equation}
\begin{equation}
     \mathcal{L}_{\text{mask}}  = \sum_r\text{BCE}(\hat{M}(r), M(r))
\end{equation}
\begin{equation}
     \mathcal{L_{\text{depth}}}= \sum_r(M(r) \otimes ||\hat{D}(r) - D(r)||_1)
\end{equation}
where $\otimes$ is element-wise production. We use both MSE and L1 loss on the rendered color for better convergence. 

In addition to encouraging the triplane to fit the object well, we still want it to be friendly to the later diffusion process. So we put several regularizations on the triplane itself. Inspired by \cite{large-vocabulary}, we use 
L2 and TV loss as follows:
\begin{equation}
\begin{aligned}
    \mathcal{L}_{\text{L2}} = & ||F||_2 \\
    \mathcal{L}_{\text{TV}} = & \sum_{1\le i < H} ||F(i, \cdot) - F(i+1, \cdot)||_1 \\+&\sum_{1\le i < W}||F(\cdot, i) - F(\cdot, i+1)||_1
\end{aligned}
\end{equation}

In order for the geometry to converge fast, we render 3 axis-aligned views in orthogonal projection and adopt these masks $O_{xy}, O_{yz}, O_{xz}$ on 3 planes to constrain where it has non-zero features, which is basically equivalent to a 3-view visual hull constraint. 
\begin{equation}
    \mathcal{L}_{\text{hull}} = ||(1 - O)\otimes F ||_1
\end{equation}
Above we neglect the subscript $xy, xz, yz$ as they are adopted on all planes. 

In general, we adopt an autodecoder architecture to fit all the triplanes. However, a large number of triplanes cannot be put into the GPU memory at the same time. So we first train the shared MLP $\mathcal{D}(\cdot;\theta)$ on a subset and fit all the 3D assets independently by freezing $\mathcal{D}(\cdot;\theta)$. The final optimization objective for either training $\mathcal{D}$ or $F$ is
\begin{equation}
\begin{aligned}
    \mathcal{L} = \mathcal{L}_{\text{color}} &+ \lambda_{\text{mask}}\mathcal{L}_{\text{mask}} + \lambda_{\text{depth}}\mathcal{L}_{\text{depth}} 
    \\
   & + \lambda_{\text{L2}}\mathcal{L}_{\text{L2}} + \lambda_{\text{TV}}\mathcal{L}_{\text{TV}} + \lambda_{\text{hull}}\mathcal{L}_{\text{hull}}
\end{aligned}
\label{eq:loss}
\end{equation}

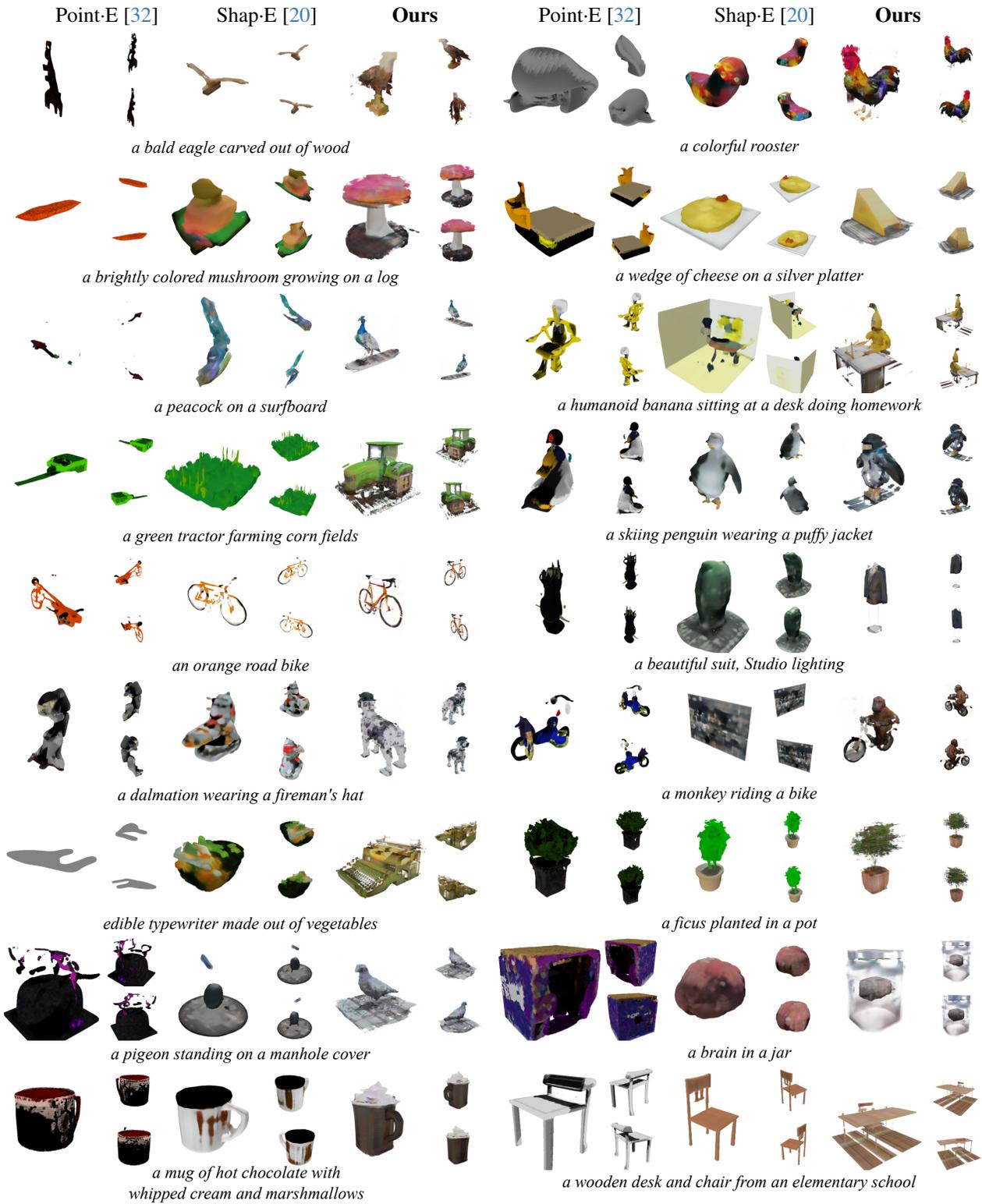
\begin{figure*}[t]
\captionsetup{justification=centering}
\centering
\scalebox{0.71}{
\begin{overpic}[scale=0.4]{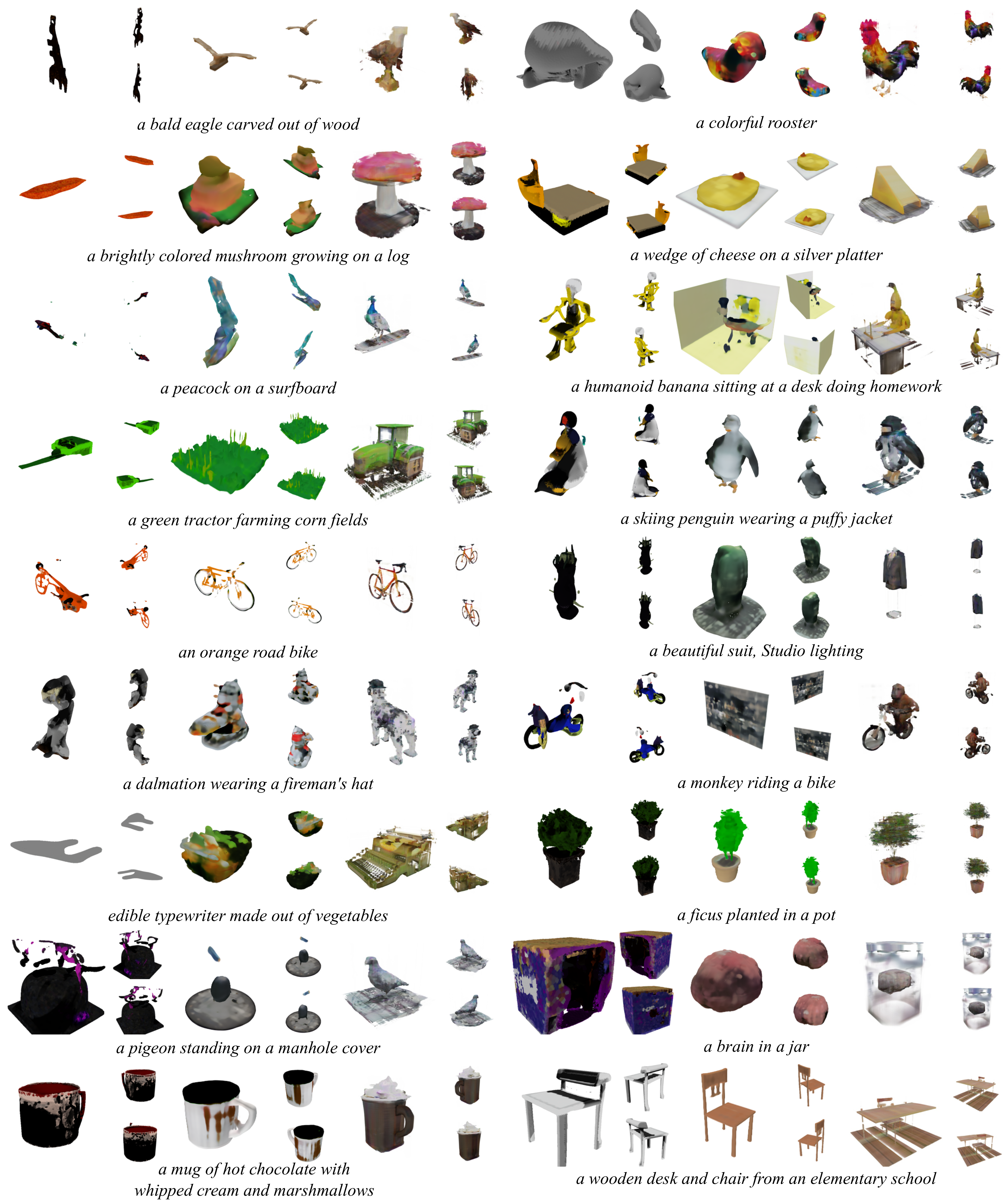}
\put(40,810){\Large Point·E~\cite{pointe}}
\put(150,810){\Large Shap·E~\cite{shape}}
\put(270,810){\Large \textbf{Ours}}

\put(370,810){\Large Point·E~\cite{pointe}}
\put(490,810){\Large Shap·E~\cite{shape}}
\put(600,810){\Large \textbf{Ours}}
\end{overpic}
}
\caption{Examples of 3D models generated by Point·E~\cite{pointe}, Shap·E~\cite{shape} and our method conditioned on various text prompts. 3D models are presented by 3 rendered views. We follow Point·E~\cite{pointe} to extract the surface for the point cloud rendering. The prompts are listed below each sample. }
\label{fig:comparison}
\end{figure*}

\begin{table*}[htbp]
    \centering
\caption{Quantitative comparisons between \sysname and state-of-the-art text-to-3D generative models on CLIP Score~\cite{clipscore} and CLIP R-Precision metrics. The R-precision is calculated based on top 1 and top 10 recall. }
    \begin{tabular}{c|ccc|ccc}
   \toprule
 \multirow{2}*{Method} & \multicolumn{3}{c|}{ViT-B-32} & \multicolumn{3}{c}{ViT-L-14} \\
  & CLIP Score & R-Precision@1 & R-Precision@10 & CLIP Score & R-Precision@1 & R-Precision@10 \\
  \midrule
 Point·E & 22.3 & 4.6 & 15.7 & 42.6 & 5.7 & 26.3 \\
 Shap·E & 62.0 & 16.0 & 41.0 & 48.3 & 17.3 & 43.7\\
 \lfi{\sysname} & \lfi{65.9} & \lfi{25.2} & \lfi{56.7} & \lfi{53.8} & \lfi{29.3} & \lfi{63.3} \\ 
  \bottomrule
    \end{tabular}
\vspace{-3mm}
\label{tab:clipscore}
\end{table*}

\subsection{Pseudo-Image Diffusion}
\label{sec:pid}
We choose Stable Diffusion~\cite{ldm} as our base model and build a pseudo-image diffusion model on top of it without altering its architecture. The two main challenges here are: (1) how to simultaneously output multiple multi-channel feature maps under the architecture, and (2) how to model the cross-plane consistency to form a reasonable underlying 3D geometry. We view each triplane representation as a set of 6 pseudo-images and stack them along the batch dimension as the model input. A global pseudo-image attention block, inspired by video and multi-view diffusion models~\cite{imagenvideo,tuneavideo,videodiffusion,mvdream}, is adapted from the original self-attention block without changing the parameters to aggregate tokens on 6 pseudo-images. 
In this way, we enrich the feature interaction across three planes and incorporate the 3D knowledge into the global attention block. So far, all the attention layers are order-invariant. Thus, to identify the input images, we allocate a learnable embedding for each pseudo-image and add it to the timestep embedding in the cross-attention block.

We directly trained the model on our fitted triplanes but observed a decline in generalization ability on 3D assets over prolonged training, particularly for prompts involving combined concepts and multiple objects, showing less robustness than image diffusion models. We believe the performance degradation is attributed to the quality and scale of 3D data. 
There are basically all single objects in the Objaverse dataset~\cite{objaverse} and still exist many misaligned text descriptions in Cap3D~\cite{cap3d}. To enhance robustness, we train the pseudo-image diffusion with additional real images from LAION-5B~\cite{laion5B} by disabling global attention blocks and learnable embedding. We observe obvious improvements in the diversity and prompt alignment after training with mixed 2D-3D data which we will discuss later in the ablation study (see \cref{sec:ablate}).

\subsection{Lightweight Refinement}
\label{sec:refine}

Upon obtaining the diffusion model, two intuitive ways to generate 3D models from input texts are considered. One can directly sample new 3D models within seconds from it and further refine it with 2D SDS. Alternatively, we can view the 3D diffusion model as a prior and jointly optimize a triplane-based NeRF with 2D and 3D SDS, like Magic123~\cite{magic123} to guarantee the 3D consistency. In practice, we find that sampling a 3D model first using our method and then refining it with only 2D diffusion models is a simple yet effective way to produce high-quality objects and guarantee 3D consistency. In the refinement, we continue to optimize the triplane generated by the pseudo-image diffusion along with the shared decoder with SDS loss. To prevent the 3D shape from being over-optimized, we refine the appearance and geometry by sampling only small noise levels in SDS. A good coarse 3D model provided by the pseudo-image diffusion gives 2D SDS a strong appearance prior, making it converge much faster than starting from scratch~\cite{dreamfusion,fantasia3D,vsd,mvdream}.

\section{Experiments}
\subsection{Implementation Details}
\paragraph{Data} For the 3D dataset, we use 20k 3D objects from Objaverse~\cite{objaverse} LVIS subset along with their corresponding captions from Cap3D~\cite{cap3d}. In the triplane fitting stage, we have standardized the size of the 3D objects by scaling them to fit within a unit-length bounding box. Subsequently, we placed them at the origin and generated 100 random views for each object at a resolution of $512\times512$.

\paragraph{Triplane fitting} 
We define the triplane size as $H = W = 256, C = 6$. The shared decoder used in volume rendering is a 4-layer MLP, with each hidden layer having a width of 32. It is trained on a set of 500 objects. Subsequently, we freeze the shared decoder and independently fit the triplane representation for each object, implementing a linearly decayed learning rate. The balancing coefficients for loss items are as follows: $\lambda_{\text{L1}} = 0.2,\lambda_{\text{mask}} = 0.1,\lambda_{\text{depth}} = 0.5,\lambda_{\text{L2}} = 0.05,\lambda_{\text{TV}} = 0.05,\lambda_{\text{hull}} = 1.0$. The triplanes are scaled to $[-1, 1]$ to ensure a similar variance with images for easy diffusion training later. The process of fitting one object on a single A100 GPU takes 2 minutes.

\paragraph{Pseudo-image diffusion} 
We start from \textit{Stable Diffusion v2.1} and train the learnable embedding with a learning rate of $10^{-3}$ while the remaining parameters are trained with $10^{-4}$ for 15,000 iterations. Training is conducted using $256\times 256$ real images from LAION-5B~\cite{laion5B} and pseudo-images with 50\% probability each. The equivalent batch size for triplanes and real images are set to 384 and 2304, respectively, to match their computational overhead. We drop the text embedding with a 5\% probability, jointly training an unconditional diffusion model to enable classifier-free guidance~\cite{cfg}.
Utilizing the v-objective~\cite{progressive} and Min-SNR strategy~\cite{minsnr} enhances convergence during diffusion training. The entire diffusion training process spans about 36 hours, utilizing 8 A100 GPUs.

\paragraph{Text-to-3D generation} In pseudo-image sampling, we use DDPM~\cite{ddpm} scheduler for 50 sample steps with a classifier free guidance~\cite{cfg} scale of 5.0. It takes 3 seconds to sample a triplane with the $256\times256$ resolution. We implement the subsequent refinement based on threestudio~\cite{threestudio}, utilizing DeepFloyd-IF~\cite{deepfloyd} as the 2D guidance model. The triplane-based NeRF is optimized for 2000 steps with a batch size of 1, typically requiring 3 minutes on an A100 GPU. The timestep $t$ is sampled in $\mathcal{U}(0.1, 0.5)$ to introduce small corrections and prevent excessive changes. The classifier-free guidance adopted in SDS is 20.

\begin{figure*}[htbp]
    \centering
    \includegraphics[width=\linewidth]{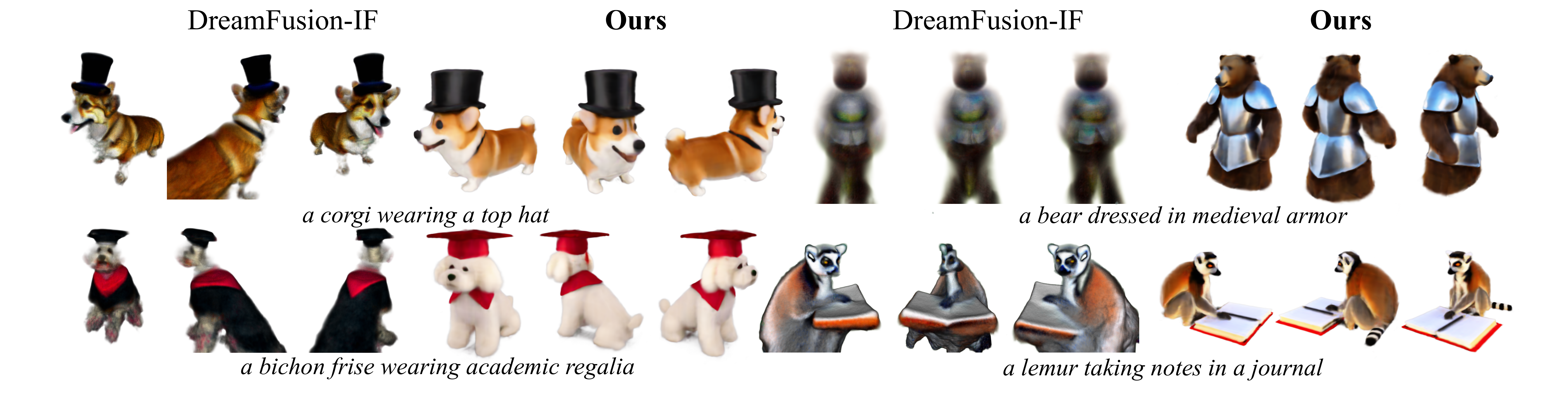}
    \vspace{-8mm}
    \caption{Examples generated by Dreamfusion-IF~\cite{dreamfusion,threestudio} and our method. \sysname can generate 3D consistent models much faster.}
    \vspace{-5mm}
    \label{fig:comp_df}
\end{figure*}

\begin{figure}[htbp]
    \centering
    \vspace{-2mm}
    \includegraphics[width=\linewidth]{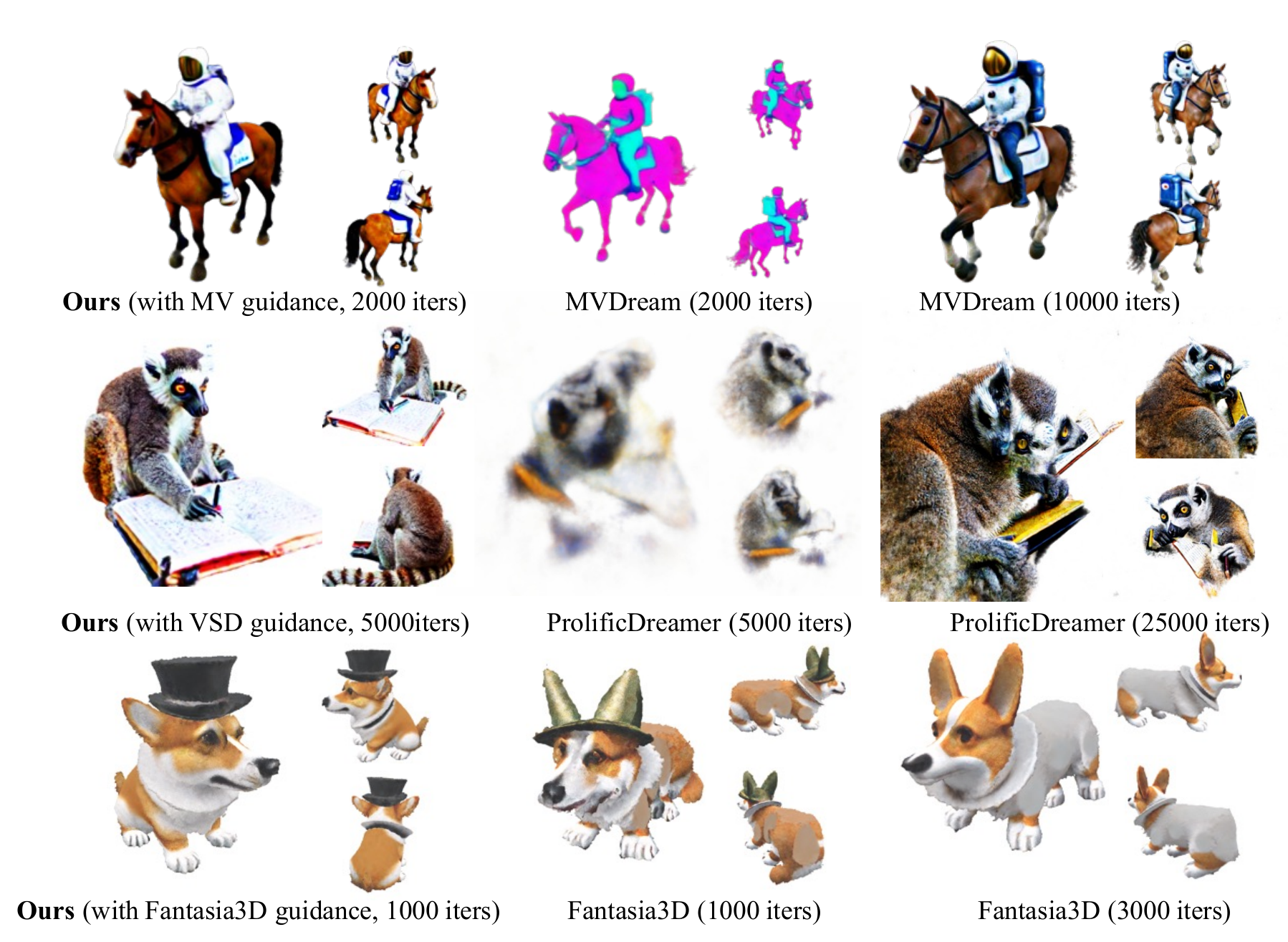}
    \vspace{-8mm}
    \caption{We replace the 2D guidance used in the refinement with those in other methods. We can achieve comparable or better results on significantly fewer iterations.}
    \label{fig:other_baseline}
    \vspace{-6mm}
\end{figure}

\subsection{Comparisons with Other Text-to-3D Methods}
We compare \sysname with general text-to-3D diffusion models, Point·E~\cite{pointe} and Shap·E~\cite{shape} and SDS-based method DreamFusion~\cite{dreamfusion}, Fantasia3D~\cite{fantasia3D}, ProlificDreamer~\cite{vsd}, MVDream~\cite{mvdream}. To evaluate the fidelity, we adhere to the metrics used in prior works~\cite{shape,dreamfusion}, namely CLIP R-Precision and CLIP similarity score~\cite{clipscore}. CLIP R-precision measures the accuracy of retrieving text prompts using rendered images based on CLIP similarity. We select 4 views at different camera poses with azimuth angles ($0, 90, 180, 270$ degrees) and the same elevation angle ($15$ degrees), performing multiple retrievals to average a score. All 415 testing prompts are from the DreamFusion gallery.

We first compare the pseudo-image diffusion model with the baseline text-to-3D diffusion models. The quantitative results are reported in \cref{tab:clipscore}. Our model significantly outperforms the other two methods in all the metrics, revealing that it could generate more text-aligned 3D models. We also demonstrate the generation quality through some qualitative cases (see \cref{fig:comparison}), each presented with 3 rendered views. The baselines, Point·E~\cite{pointe} and Shap·E~\cite{shape}, can generate simple objects that are part of complex prompts but struggle to capture the semantics of intricate text prompts, for example, depicting ``corn fields" without the tractor or a ``penguin" without the skiing activity. 
We believe the limited generalization ability is partly because they are both trained only on 3D datasets with variable quality of text annotation and 3D geometry. However, leveraging the similarity between pseudo-images and 2D real images, our method can learn rich concepts from limited 3D data and achieve reasonable generation quality even with prompts involving concept combination.

\begin{figure}[htbp]
    \centering
    \vspace{-2mm}
    \includegraphics[width=0.6\linewidth]{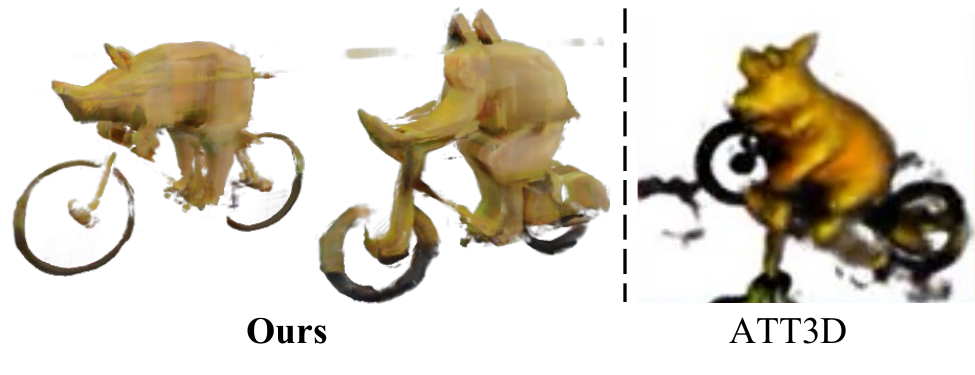}
    \vspace{-5mm}
    \caption{Our diffusion model beats ATT3D in terms of 3D consistency and can produce diverse results.}
    \vspace{-5mm}
    \label{fig:att3d_main}
\end{figure}

We also compare the refinement stage with  SDS-based methods on the generation efficiency and quality. Several examples by DreamFusion-IF from threestudio~\cite{threestudio} and our method are compared in \cref{fig:comp_df}. The baseline method requires tons of iterations in hours while still performing poorly in 3D consistency and convergence. Our method can generate 3D consistent models and converge much faster with an even smaller batch size in only 3 minutes by leveraging a good 3D initialization sampled from the diffusion model. Also, our method is compatible with other SDS variants. As demonstrated in \cref{fig:other_baseline}, we compare our method with other SDS-based methods by switching the 2D guidance to the corresponding variants. We can achieve superior or comparable quality with even much faster convergence. In terms of efficient generation, ATT3D~\cite{att3d} has an inference runtime of 1s, much faster compared to optimization-based methods. But we contend that it is much less generalizable than our method. Limited by a prompt set, it can only generate new shapes by combining seen words while \sysname supports complex concepts. Our sampling takes around 3s and ATT3D has a runtime of 1s. As it is not open-sourced, we show the results from their paper in \cref{fig:att3d_main}. Furthermore, it employs a deterministic mapping network with no diversity.


\begin{table*}[t]
\caption{Quantitative results of the ablation experiments on probability for training with 2D data and classifier free guidance.}
\label{tab:ablation_hyp}
    \centering
    \begin{tabular}{cc|ccc|ccc}
   \toprule
 \multirow{2}*{CFG} & \multirow{2}*{2D\%} & \multicolumn{3}{c|}{ViT-B-32} & \multicolumn{3}{c}{ViT-L-14} \\
  & & CLIP Score & R-Precision@1 & R-Precision@10 & CLIP Score & R-Precision@1 & R-Precision@10 \\
  \midrule
 1.0 & 0.5 & 59.2 & 8.4 & 26.9 & 45.6 & 10.1 & 28.0 \\
 3.0 & 0.5 & 65.9 & 24.6 & 55.1 & 53.2 & 29.1 & 61.7 \\
 5.0 & 0.5 & 65.9 & \lfi{25.2} & \lfi{56.7} & \lfi{53.8} & 29.3 & 63.6\\
 7.5 & 0.5 & \lfi{66.0} & 24.4 & \lfi{56.7} & \lfi{53.8} & 29.3 & 63\\
 10.0 & 0.5 & 64.4 & 23.4 & 54.2 & 53.2 & 28.1 & 61.6 \\
 \midrule
 5.0 & 0 & 61.5 & 15.7 & 41.8 & 47.4 & 16.7 &  43.1 \\
 5.0 & 0.3 & 64.7 & 22.2  & 53.9 & 51.5 & 23.9  & 57.3 \\
 5.0 & 0.7 & 65.4 & \lfi{25.2}  & 56.3 & 53.4 & \lfi{30.8}  & \lfi{64.0} \\
  \bottomrule
    \end{tabular}
    \vspace{-3mm}
\end{table*}

\begin{figure}[htbp]
    \centering
    \vspace{2mm}
    \scalebox{0.7}{
    \begin{overpic}[scale=0.8]{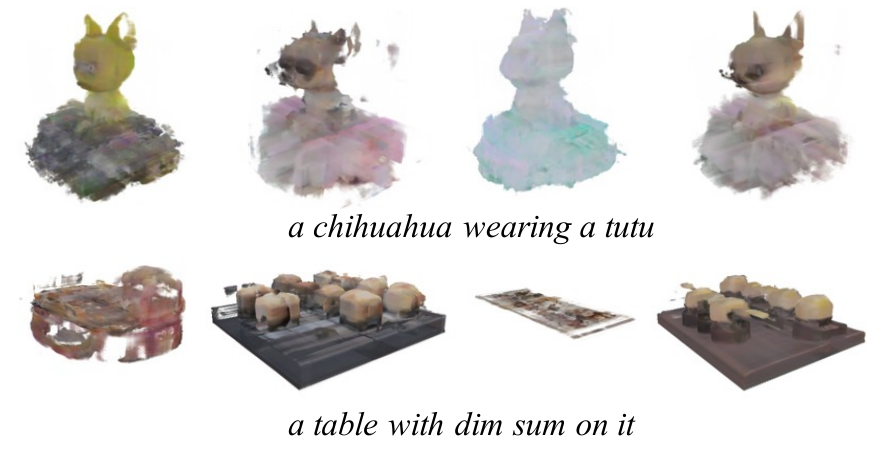}{
    \put(20,180){CFG = 1.0}
    \put(105,180){CFG = 15.0}
    \put(190,180){2D\%=0.0}
    \put(285,180){\textbf{Ours}}
    }
    \end{overpic}
    }
    \vspace{-8mm}
\caption{Qualitative results of ablation experiments on different classifier-free guidance scales and the probabilities of training with real images.}
\vspace{-5mm}
\label{fig:ablate-hyp}
\end{figure}
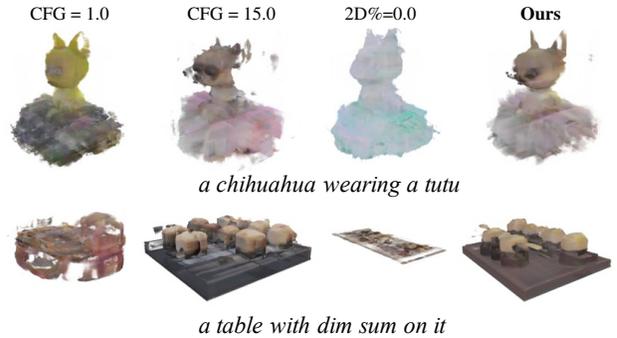

\subsection{Ablation study}

\begin{figure}[htbp]
    \centering
    \scalebox{0.9}{
    \begin{overpic}[scale=0.4]{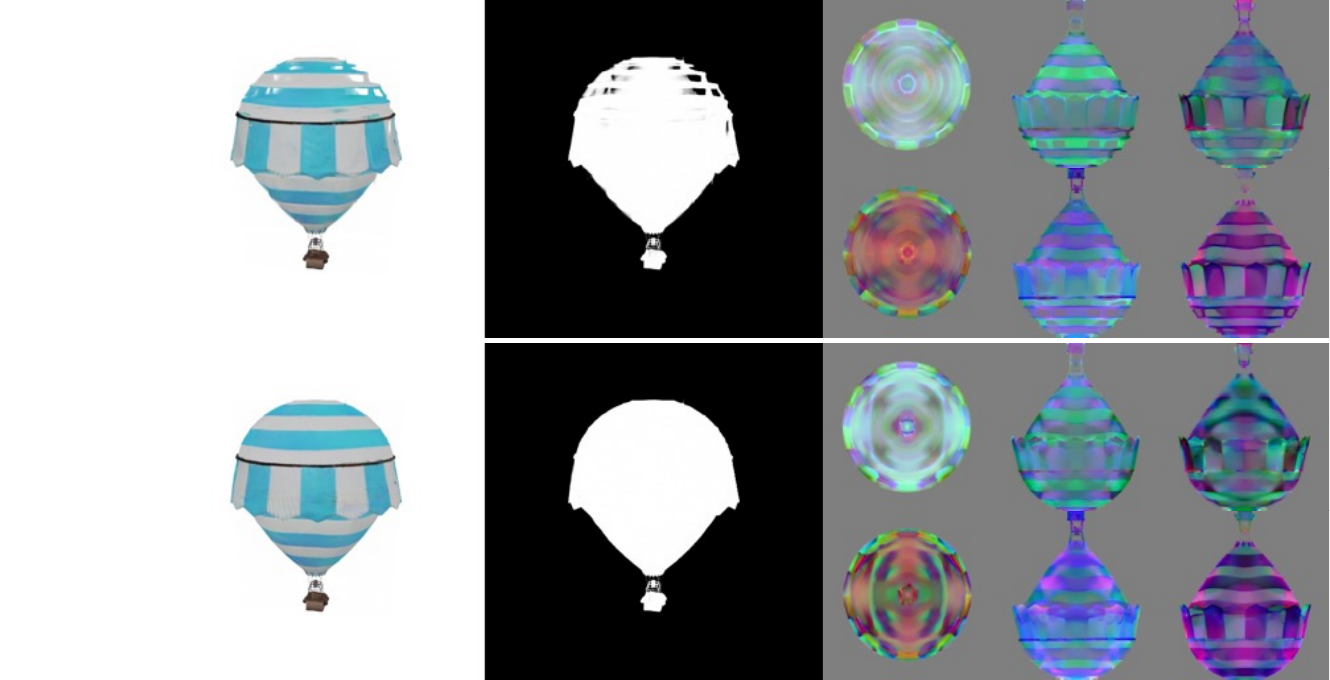}
    \put(0,100){w/o $\mathcal{L}_{\text{depth}}$}
    \put(0,30){w/ $\mathcal{L}_{\text{depth}}$}
    \end{overpic}}
    \caption{Fitting the object w/ and w/o the depth loss. Rendered RGB images, opacity images, and pseudo-images are shown.}
    \label{fig:ablation-depth}
    \vspace{-6mm}
\end{figure}

\paragraph{Depth loss in triplane fitting}
To enforce a correct underlying geometry in the triplane fitting stage, we incorporate a depth loss on the expected depth in volume rendering. 
We compare the outcomes of using depth loss versus fitting without it, as illustrated in \cref{fig:ablation-depth}. It is evident that the imposition of the depth loss constraint is advantageous for forming the correct geometry surface. In the absence of depth loss, the object's surface tends to exhibit cavities near the spherical area, resulting in hollowness in the corresponding part of the triplane. Therefore, adopting depth loss during the fitting stage is crucial to ensure a plausible triplane distribution. Both examples utilize the same shared decoder and hyper-parameters, except for $\lambda_{\text{depth}}$.

\paragraph{Probability of training with real images}
\label{sec:ablate}
In the pseudo-image diffusion training phase, we observe that jointly fine-tuning the model with both pseudo-images and real images is effective in preserving the prior knowledge from the 2D pre-trained diffusion model. Consequently, the generalization ability can be maintained to some degree. However, using different probabilities to train with real images can lead to distinct outcomes, such as an excessive focus on 2D image generation, neglecting triplane quality, or a susceptibility to overfitting to 3D datasets.  We evaluate various training settings using the metrics and prompt datasets used above. The results are presented in \cref{tab:ablation_hyp}, where the probability of utilizing real images is marked as 2D\%. When training with only 3D data, the model behaves similarly to Shap·E~\cite{shape}. They both lose the rich semantic knowledge in the 2D data and are prone to overfitting to low-quality 3D data, resulting in poor performance on more generalized complex prompts. The generalization ability of the model increases significantly as the proportion of 2D data increases, and models trained with real images at probabilities of 50\% and 70\% perform comparably. We select 50\% as our final model. Additionally, we show several qualitative results in \cref{fig:ablate-hyp}. Training without real images makes the model less generalizable and unable to comprehend complex text inputs.

\paragraph{Classifier free guidance} 
We experiment with various classifier-free guidance scales to identify the optimal match between the generated 3D content and the input text prompt. The qualitative results for different scales are presented in \cref{tab:ablation_hyp}. It is evident that \sysname performs best when the scale is in the range of $5.0\sim7.5$. We attribute this to the parallel training with real images, aligning the optimal range with sampling the image diffusion model. The qualitative results in \cref{fig:ablate-hyp} indicate that the 3D model poorly conforms to the text and exhibits rough geometry when the scale is too small, while a large scale introduces noisy geometry although the semantics are more text-aligned.

\section{Limitations}
Although \sysname demonstrates the ability to efficiently generate 3D consistent shapes from text prompts, there still exist some limitations left for the future to address. First, our training procedure involves a per-object triplane fitting phase, which leads to a linear increase in training costs when scaling up to larger 3D datasets. This might be solved by encoding triplanes from point clouds or multi-view images with an autoencoder architecture like \cite{pointe,3dgen}. Additionally, the capacity of our scene representation is limited by the feature dimensions. When it comes to more detailed 3D generation, higher dimensional features are inevitably required, which raises the number of pseudo-images and poses a challenge in modeling the relationships between them. This may be solved by designing more explicit and effective feature interactions.

\section{Conclusion}
In this paper, we introduce \sysname, a novel and efficient framework that utilizes the pre-trained text-to-image diffusion models to generate high-quality 3D shapes in minutes. On the one hand, it fine-tunes a pre-trained 2D diffusion model into a 3D diffusion model, enabling both 3D generative capabilities and generalization derived from the 2D model. On the other, it utilizes score distillation sampling of 2D diffusion models to quickly improve the quality of the sampled 3D shapes. \sysname enables the migration of knowledge from image to triplane generation by treating it as a set of pseudo-images. We adapt the modules in the pre-training model to enable hybrid training using pseudo and real images, which has proved to be a well-established strategy for improving generalizability. The efficiency of \sysname is highlighted by its ability to sample diverse 3D models in seconds and refine them in minutes. The experimental results confirm the advantages of \sysname over existing methods based on either 3D diffusion models or lifting 2D diffusion models in terms of fast generation of 3D consistent and high-quality models. The proposed \sysname stands as a promising advancement in the field of text-to-3D generation, and we hope it will inspire more research into 3D generation leveraging the knowledge in both 2D and 3D data.

\noindent\textbf{Acknowledgement} This work was supported by the National Key Research and Development Program of China (No. 2023YFF0905104),
the Natural Science Foundation of China (No. 62132012, No. 62361146854), Beijing Municipal Science and Technology Project (No. Z221100007722001) and Tsinghua-Tencent Joint Laboratory for Internet Innovation Technology.

{
    \small
    \bibliographystyle{ieeenat_fullname}
    \bibliography{main}
}


\end{document}